\documentclass[runningheads]{llncs}
\usepackage{graphicx}
\usepackage{amsmath,amssymb} % define this before the line numbering.
\usepackage{color}
\usepackage[width=122mm,left=12mm,paperwidth=146mm,height=193mm,top=12mm,paperheight=217mm]{geometry}
\usepackage{bm}
\usepackage{booktabs}
\usepackage{subfigure}
\usepackage{tabularx}
\usepackage{booktabs}
\usepackage{multirow}
\usepackage{epsfig}
\begin{document}
% \renewcommand\thelinenumber{\color[rgb]{0.2,0.5,0.8}\normalfont\sffamily\scriptsize\arabic{linenumber}\color[rgb]{0,0,0}}
% \renewcommand\makeLineNumber {\hss\thelinenumber\ \hspace{6mm} \rlap{\hskip\textwidth\ \hspace{6.5mm}\thelinenumber}}
% \linenumbers
\pagestyle{headings}
\mainmatter

\title{Deep Attributes Driven Multi-Camera Person Re-identification} % Replace with your title

\titlerunning{title running}

\authorrunning{authors running}

\author {Chi Su$^1$, Shiliang Zhang$^1$, Junliang Xing$^2$, Wen Gao$^1$ and Qi Tian$^3$}%{Anonymous ECCV submission}
\institute{$^1$Peking University, Beijing, China.\\
           $^2$Chinese Academy of Sciences, Beijing, China\\
           $^3$Department of Computer Science, University of Texas at San Antonio, USA\\
\email{\{chisu, slzhang.jdl, wgao\}@pku.edu.cn  jlxing@nlpr.ia.ac.cn qi.tian@utsa.edu}
}

%Please write out author names in full in the paper, i.e. full given and family names.
%If any authors have names that can be parsed into FirstName LastName in multiple ways, please include the correct parsing, in a comment to the volume editors:
%\index{Lastnames, Firstnames}
%(Do not uncomment it, because you may introduce extra index items if you do that...)

%\institute{Institute}

\maketitle

\begin{abstract}
The visual appearance of a person is easily affected by many factors like pose variations, viewpoint changes and camera parameter differences. This makes person Re-Identification (ReID) among multiple cameras a very challenging task. This work is motivated to learn mid-level human attributes which are robust to such visual appearance variations. And we propose a semi-supervised attribute learning framework which progressively boosts the accuracy of attributes only using a limited number of labeled data. Specifically, this framework involves a three-stage training. A deep Convolutional Neural Network (dCNN) is first trained on an independent dataset labeled with attributes. Then it is fine-tuned on another dataset only labeled with person IDs using our defined triplet loss. Finally, the updated dCNN predicts attribute labels for the target dataset, which is combined with the independent dataset for the final round of fine-tuning. The predicted attributes, namely \emph{deep attributes} exhibit superior generalization ability across different datasets. By directly using the deep attributes with simple Cosine distance, we have obtained surprisingly good accuracy on four person ReID datasets. Experiments also show that a simple distance metric learning modular further boosts our method, making it significantly outperform many recent works.

\end{abstract}

\section{Introduction} \label{sec:introduction}
\begin{figure}[t]
  \centering
  \includegraphics[width=0.65\linewidth]{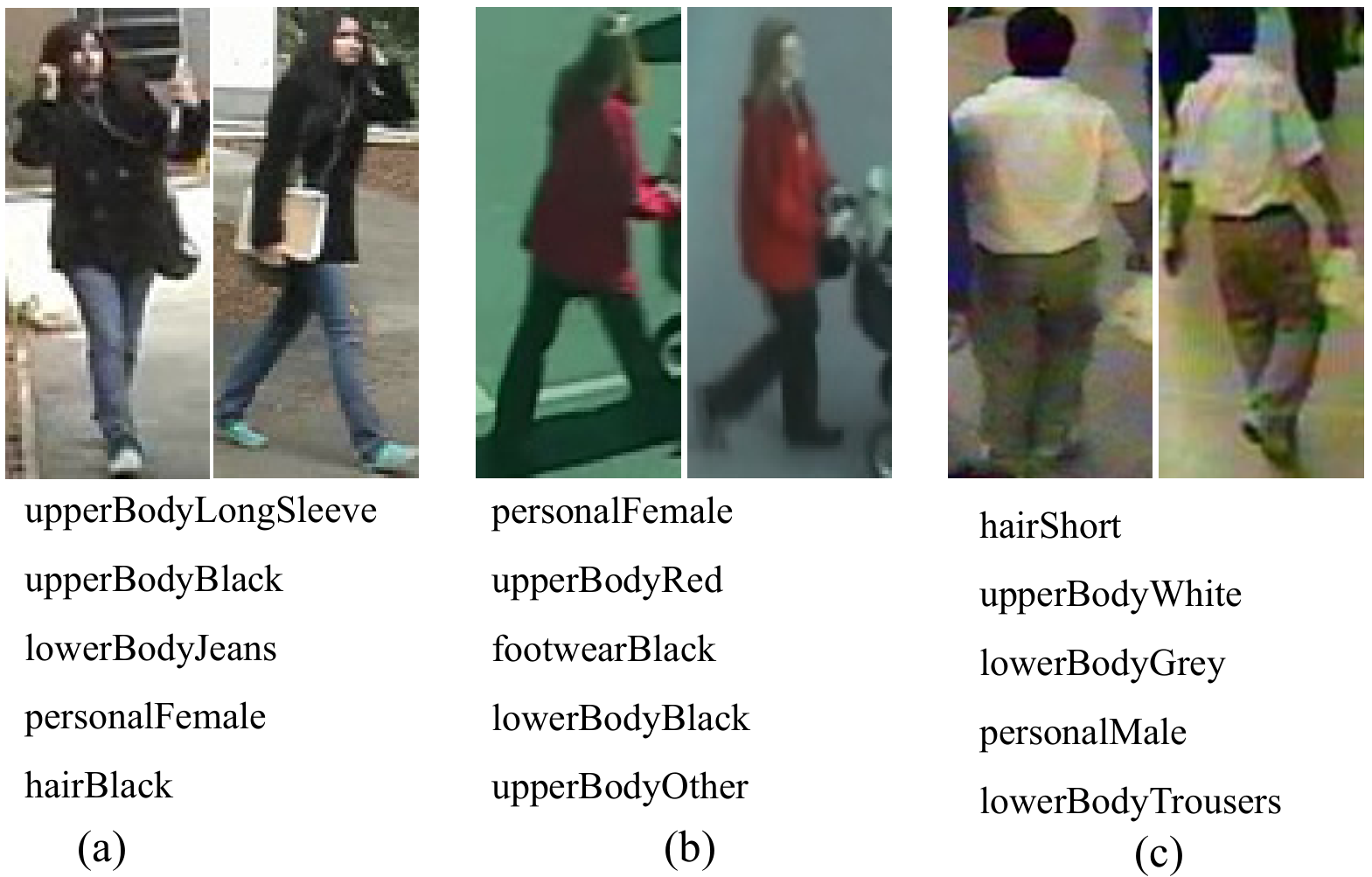}\\
  \caption{Example images of the same person taken by two cameras from three datasets: (a) \emph{VIPeR}~\cite{gray2007evaluating}, (b) \emph{PRID}~\cite{hirzer2011person}, and (c) \emph{GRID}\cite{loy2013person}. This figure also shows five of our predicted attributes shared by these two images.
}
  \label{fig:example}
\end{figure}

Person Re-Identification (ReID) targets to identify the same person from different cameras, datasets, or time stamps. As illustrated in Fig.~\ref{fig:example}, factors like viewpoint variations, illumination conditions, camera parameter differences, as well as body pose changes make person ReID a very challenging task. Due to its important applications in public security, \emph{e.g.}, cross camera pedestrian searching, tracking, and event detection, person ReID has attracted lots of attention from both the academic and industrial communities. Currently, research on this topic mainly focus on two aspects: a) extracting and coding local invariant features to represent the visual appearance of a person~\cite{farenzena2010person,cheng2011custom,ma2012bicov,liu2012person,zhao2013unsupervised,wang2014person,Zheng_2015_CVPR} and b) learning a discriminative distance metric hence the distance of features from the same person can be smaller~\cite{ma2013domain,dikmen2011pedestrian,hirzer2012relaxed,pedagadi2013local,yan2007graph,kostinger2012large,xiong2014person,liu2013pop,li2013learning,zheng2013re,wang2014personvideo,chen2015similarity,liao2015person,chen2015mirror,shen2015person,liao2015efficient,ding2015deep,pengunsupervised}.

Although significant progress has been made from previous studies, person ReID methods are still not mature enough for real applications. Local features mostly describe the low-level visual appearance, hence are not robust to variances of viewpoints, body poses, \emph{etc}. On the other side, distance metric learning suffers from the poor generalization ability and the quadratic computational complexity, \emph{e.g.}, different datasets present different visual characteristics corresponding to different metrics.
Compared with low-level visual feature, human attributes like long hair, blue shirt, \emph{etc}., represent mid-level semantics of a person. As illustrated in Fig.~\ref{fig:example}, attributes are more consistent for the same person and are more robust to the above mentioned variances. Some recent works hence have started to use attributes for person ReID~\cite{layne2012person,layne2012towards,layne2014attributes,layne2014re,su2015tracklet,su2015multi}. Because human attributes are expensive for manual annotation, it is difficult to acquire enough training data for a large set of attributes. This limits the performance of current attribute features. Consequently, low-level visual features still play a key role and attributes are mostly used as auxiliary features~\cite{layne2014attributes,layne2014re,su2015tracklet,su2015multi}.

Recently, deep learning has exhibited promising performance and generalization ability in various visual tasks. For example in~\cite{Krizhevsky:NIPS12}, an eight-layer deep Convolutional Neural Network (dCNN) is trained with large-scale images for visual classification. The modified versions of this network also perform impressively in object detection~\cite{Girshick:PAMI15} and segmentation~\cite{Long:CVPR14}. Motivated by the issues of low level visual features and the success of dCNN, our work targets to learn a dCNN to detect a large set of human attributes discriminative enough for person ReID. Due to the diversity and complexity of human attributes, it is a laborious task to manually label enough of attributes for dCNN training. The key issues are hence how to train this dCNN from a partially-labeled dataset and ensure its discriminative power and generalization ability in the person ReID tasks.
\begin{figure}[t]
  \centering
  \includegraphics[width=0.80\linewidth]{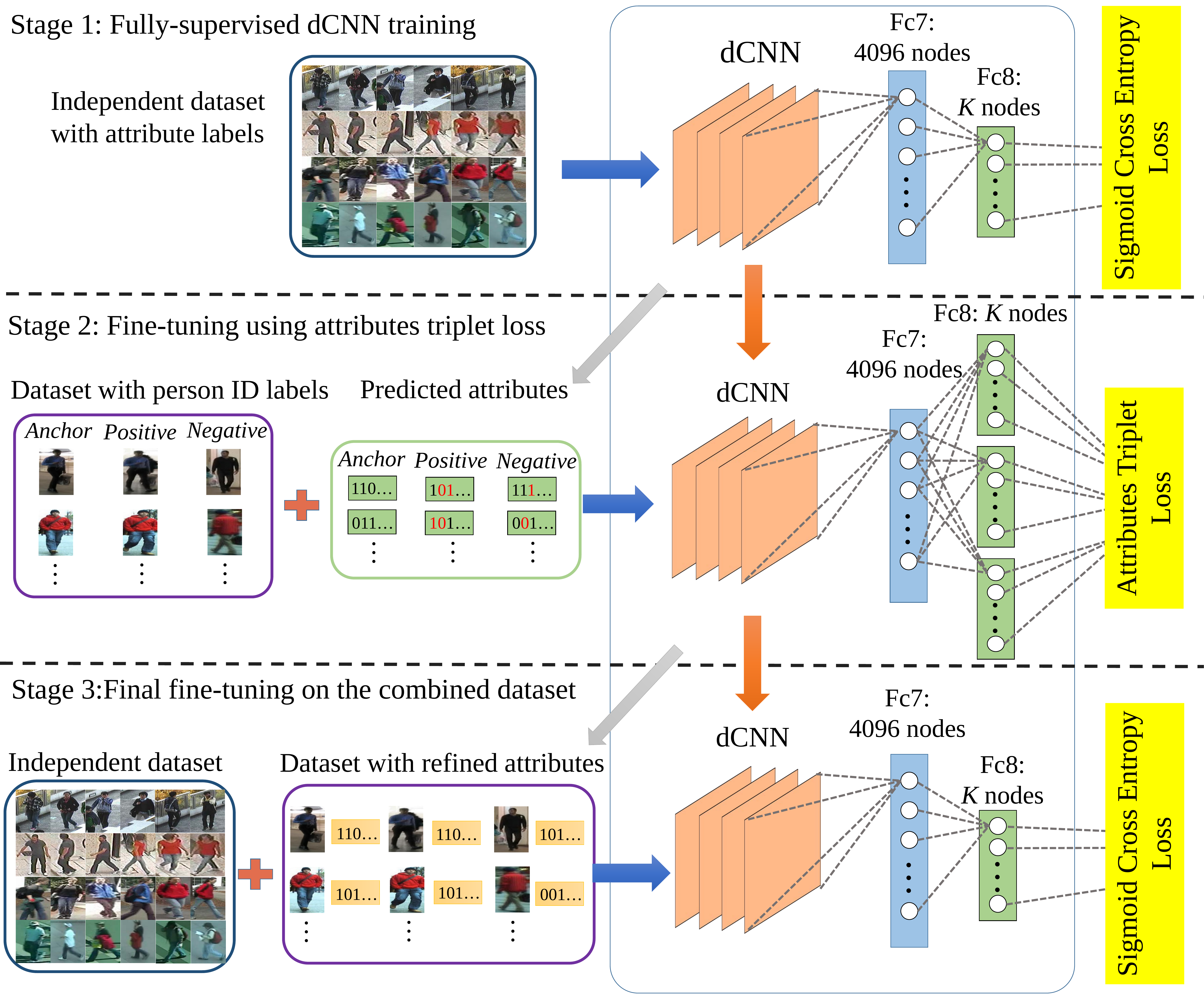}\\

  \caption{Illustration of Semi-supervised Deep Attribute Learning (SSDAL).}

  \label{fig:framework}

\end{figure}

To address these issues, we propose a Semi-supervised Deep Attribute Learning (SSDAL) algorithm. As illustrated in Fig.~\ref{fig:framework}, this algorithm involves three stages. The first stage uses an independent dataset with attribute labels to perform fully-supervised dCNN training. The resulting dCNN produces initial attribute labels for the target dataset. To improve the discriminative power of these attributes for ReID task, we start the second stage of training, \emph{i.e.}, fine-tuning the network using the person ID labels and our defined \emph{attributes triplet loss}. The training data for fine-tuning can be easily collected because the person ID labels are readily accessible in many person tracking datasets. The attributes triplet loss updates the network to enforce that the same person has more similar attributes and vice versa. This fine-tuned dCNN hence predicts initial attribute labels for target datasets. Finally in the third stage, the initially labeled target dataset plus the original independent dataset are combined for the final stage of fine-tuning. The attributes predicted by the final dCNN model are named as \emph{deep attributes}. In this manner, the dCNN is firstly trained with the independent dataset, then is refined to acquire more discriminative power for person ReID task. Because this procedure involves one dataset with attribute labels and another without attribute labels, we call it a semi-supervised learning.

To validate the performance of deep attributes, we test them on four popular person ReID datasets \emph{without} combining with the local visual features. The experimental results show that deep attributes perform impressively, \emph{e.g.}, they significantly outperform many recent works combining both attributes and local features~\cite{layne2014attributes,layne2014re,su2015tracklet,su2015multi}. Note that, predicting and matching deep attributes make person ReID system significantly faster, because it no longer needs to extract and code local features, compute distance metric, and fuse with other features.

Our contributions can be summarized as follows: 1) we propose a three-stage semi-supervised deep attribute learning algorithm, which makes learning a large set of human attributes from a limited number of labeled attribute data possible, 2) deep attributes achieve promising performance and generalization ability on four person ReID datasets, and 3) deep attributes release the previous dependencies on local features, thus make the person ReID system more robust and efficient. To the best of our knowledge, this is an original work predicting human attributes using dCNN for person ReID tasks. The promising results of this work guarantees further investigation in this direction.

%The remainder of this paper is organized as follows. Section~\ref{sec:related_work} reviews the related work on person ReID. Section~\ref{sec:proposed_approach} presents our approach for CNN training. Section~\ref{sec:experiment} analyzes experimental results, followed by the conclusions in Section~\ref{sec:conclusions}.

  %The first stage trains an initial dCNN on an independent dataset labeled with attributes. The second stage fine-tunes the initial dCNN based on our attributes triplet loss on a larger dataset annotated with person IDs. The refined dCNN hence labels the target dataset by predicting refined attribute labels. Finally, the independent dataset plus the labeled target dataset are combined for the final stage of fine-tuning.
\section{Related Work} \label{sec:related_work}
This work learns a dCNN for attribute prediction and person ReID. It is closely related to works using deep learning for attribute prediction and person ReID.

Currently, many studies have applied deep learning to attributes learning~\cite{shankar2015deep,chen2015deep}. Shankar \emph{et al.}~\cite{shankar2015deep} propose a deep-carving neural net to learn attributes for natural scene images. Chen \emph{et al.}~\cite{chen2015deep} use a double-path deep domain adaptation network to get the fine-grained clothing attributes. Our work differs from them in the aspects of motivation and methodology. We are motivated by how to learn attributes of the human cropped from surveillance videos from a small set of data labeled with attributes. Our semi-supervised learning framework consistently boosts the discriminative power of dCNN and attributes for person ReID.

Inspired by the promising performance of deep learning, some researchers begin to use deep learning to learn visual features and distance metrics for person ReID~\cite{li2014deepreid,yi2014deep,ahmed5improved,ding2015deep}. In \cite{li2014deepreid}, Li \emph{et al.} use a deep filter pairing neural network for person ReID, where two paired filters of two cameras are used to automatically learn optimal features. In \cite{yi2014deep}, Yi \emph{et al.} present a ``siamese'' convolutional network for deep distance metric learning. In \cite{ahmed5improved}, Ahmed \emph{et al.} devise a deep neural network structure to transform person re-identification into a problem of binary classification, which judges whether a pair of images from two cameras is the same person. In \cite{ding2015deep}, Ding \emph{et al.} present a scalable distance learning framework based on the deep neural network with the triplet loss. Despite of their efforts to find better visual features and distance metrics, the above mentioned works are designed specifically for certain datasets and are dependent on their camera settings. Differently, we use deep learning to acquire general camera-independent mid-level representations. As a result, our algorithm shows better flexibility, \emph{e.g.}, it could handle person ReID tasks on datasets containing different number of cameras.

Some recent works also use triplet loss for person ReID~\cite{chen2015similarity,paisitkriangkrai2015learning}. Our work uses attributes triplet loss for dCNN fine-tuning. This differs from the goals in these works, \emph{i.e.}, learning distance metric among low-level features. Therefore, these works also suffer from the low flexility and the quadratic complexity.

\section{Proposed Approach} \label{sec:proposed_approach}

\subsection{Framework}

Our goal is to learn a large set of human attributes for person ReID through dCNN training. We define $A = \{a_1,a_2,...,a_K\}$ as an attribute label containing $K$ attributes, where $a_i\in\{0,1\}$ is the binary indicator of the $i$-th attribute. Our goal is hence learning an attribute detector $\mathcal{O}$, which predicts the attribute label $A_I$ for any input image $I$, \emph{i.e.},
\begin{equation}
\begin{aligned}
A_I = \mathcal{O}(I).
\label{eq:att_detect}
\end{aligned}
\end{equation}

Because of the promising discriminative power and generalization ability, we use dCNN model as the detector $\mathcal{O}(\cdot)$. However, dCNN training requires large-scale training data labeled with human attributes. Manually collecting such data is also too expensive to conduct. To ensure effective learning of a dCNN model for person ReID from only a small amount of labeled training data, we propose the Semi-supervised Deep Attribute Learning (SSDAL) algorithm.

As illustrated in Fig.~\ref{fig:framework}, the basic idea of SSDAL is firstly training an initial dCNN on an independent dataset labeled with attributes. The limited scale and label accuracy of the independent dataset motivate us to introduce the second stage of training, which utilizes the easily acquired person ID labels to refine the initial dCNN. The updated dCNN hence initially labels the target dataset by predicting attribute labels. Finally, the independent dataset plus the initially labeled target dataset are combined for the final stage of fine-tuning. In the followings, we introduce the three stages of training in detail.

\subsection{Fully-Supervised dCNN Training}

We define the independent training set with attribute labels as $T=\{t_1,t_2,...,t_N\}$, where $N$ is the number of samples. In $T$, each sample is labeled with a binary attribute label, \emph{e.g.}, the label of the $n$-th instance $t_n$ is $A_n$.

In the first stage of training, we use $T$ as the training set for fully-supervised learning. We refer to the AlexNet~\cite{Krizhevsky:NIPS12} to build our dCNN model for its promising performance in various vision tasks. Specifically, our dCNN is also a 8-layer network, including 5 convolutional layers and 3 fully connected layers, where the 3rd fully connected layer predicts the attribute labels. The kernel and filter sizes of each layer in our architecture are the same with the ones in~\cite{Krizhevsky:NIPS12,shankar2015deep}. The only difference with AlexNet is that we use a sigmoid cross-entropy loss layer instead of the softmax loss layer for its better performance in multi-label prediction. We denote the dCNN model learned in this stage as $\mathcal{O}^{S1}$.
$\mathcal{O}^{S1}$ could predict attribute labels for any test sample. However, as illustrated in our experiments, the discriminative power of $\mathcal{O}^{S1}$ is weak because of the limited scale and label accuracy of the independent training set. We proceed to introduce our second stage of training.

\subsection{dCNN Fine-tuning with Attributes Triplet Loss}

In the second stage, a larger dataset is used to fine tune the previous dCNN model $\mathcal{O}^{S1}$. The goal of our dCNN model is predicting attribute labels for person ReID tasks. The predicted attribute labels thus should be similar for the same person. Motivated by this, we use person ID labels to fine-tune $\mathcal{O}^{S1}$ and produce similar attribute labels for the same person and vice versa. We denote the dataset with person ID labels as $U=\{u_1,u_2,...,u_M\}$, where $M$ is the number of samples and each sample has a person ID label $l$, \emph{e.g.}, the $m$-th instance $u_m$ has person ID $l_m$.

In the second stage of training, we first use $\mathcal{O}^{S1}$ to predict the attribute label $\tilde{A}$ of each sample in $U$. For the attribute label $\tilde{A_m}$ of the $m$-th sample, we set the indicators of attributes with top $p$ highest confidence scores as 1 and set the others as 0. Note that, $p$ can be selected according to the average number of positive attributes in person ReID tasks. It is experimentally set as 10 in this paper. After this, we use the person ID labels to measure the annotation errors of $\mathcal{O}^{S1}$.

The annotation error of the $\mathcal{O}^{S1}$ is computed among three samples. The three samples are randomly selected from the $U$ through the following steps: 1) select an \emph{anchor} sample $u_{(a)}$, 2) select another \emph{positive} sample $u_{(p)}$ with the same person ID with $u_{(a)}$, and 3) select a \emph{negative} sample $u_{{n}}$ with different person ID. Thus, a triplet $[u_{(a)}, u_{(p)}, u_{(n)}]$ is constructed, where the subscripts $(a)$, $(p)$, and $(n)$ denote $anchor$, $positive$, and $negative$ samples, respectively. The attributes of the $e$-th triplet predicted by $\mathcal{O}^{S1}$ are $\tilde{A}_{(a)}^{(e)}$, $\tilde{A}_{(p)}^{(e)}$, and $\tilde{A}_{(n)}^{(e)}$ at the beginning of the fine-tuning, respectively.

The objectives of the fine-tuning is minimizing the triplet loss through updating the $\mathcal{O}^{S1}$, \emph{i.e.}, minimize the distance between the attributes of $u_{(a)}$ and $u_{(p)}$, meanwhile maximize the distance between $u_{(a)}$ and $u_{(n)}$. We call this triplet loss as attributes triplet loss. We hence could formulate our objective function for fine-tuning as:
\begin{equation}
\begin{aligned}
\mathbf{D}&\left(A_{(a)}^{(e)}, A_{(p)}^{(e)}\right) + \theta  < \mathbf{D}\left(A_{(a)}^{(e)},A_{(n)}^{(e)}\right),\\
&\forall\left(A_{(a)}^{(e)}, A_{(p)}^{(e)}, A_{(n)}^{(e)}\right) \in \mathcal{T},
\end{aligned}
\label{eq:triplet0}
\end{equation}
where $\mathbf{D}(.)$ represents the distance function of the two binary attribute vectors, $A_{(a)}^{(e)}$, $A_{(p)}^{(e)}$ and $A_{(n)}^{(e)}$ are predicted attributes of the $e$-th triplet during the fine-tuning. Then, the corresponding loss function can be formulated as:
\begin{equation}
\begin{aligned}
\mathcal{L}=\sum_e^E\max\biggl(&0, \mathbf{D}\left(A_{(a)}^{(e)}, A_{(p)}^{(e)}\right)+\theta-\mathbf{D}\left(A_{(a)}^{(e)},A_{(n)}^{(e)}\right)\biggr),
\end{aligned}
\label{eq:tripletloss}
\end{equation}
where $E$ represents the number of triplets. In Eq.~\eqref{eq:tripletloss}, if the $\mathbf{D}\left(A_{(a)}^{(e)},A_{(n)}^{(e)}\right)-\mathbf{D}\left(A_{(a)}^{(e)}, A_{(p)}^{(e)}\right)$ is larger than $\theta$, the loss would be zero. Therefore, parameter $\theta$ largely controls the strictness of the loss.

The above loss function essentially enforces the dCNN to produce similar attributes for the same person. However, the person ID label is not strong enough to train the dCNN with accurate attributes. Without proper constraints, the above loss function may generate meaningless attribute labels and easily over-fit the training dataset $U$. For example, imposing a large number meaningless attributes to two samples of a person may decrease the distance between their attribute labels, but does not help to improve the discriminative power of the dCNN. Therefore, we add several regularization terms and modify the original loss function as:
\begin{equation}
\begin{aligned}
\mathcal{L}=\sum_e^E\Biggl\{\max\biggl(&0, \mathbf{D}\left(A_{(a)}^{(e)}, A_{(p)}^{(e)}\right)+\theta-
&\mathbf{D}\left(A_{(a)}^{(e)},A_{(n)}^{(e)}\right)\biggr) + \gamma \times \mathcal{E}\Biggr\}£¬
\end{aligned}
\label{eq:ourloss}
\end{equation}
\begin{equation}
\begin{aligned}
\mathcal{E} = \mathbf{D}\bigl(A_{(a)}^{(e)},\tilde{A}_{(a)}^{(e)}\bigr)+\mathbf{D}\bigl(A_{(p)}^{(e)},\tilde{A}_{(p)}^{(e)}\bigr) + \mathbf{D}\bigl(A_{(n)}^{(e)},\tilde{A}_{(n)}^{(e)}\bigr),
\end{aligned}
\end{equation}
%where $\tilde{A}$ represents the predicted attribute label at the beginning of the fine-tunning.
where $\mathcal{E}$ denotes the amount of change in attributes caused by the fine-tuning. The loss in Eq.~\eqref{eq:ourloss} not only ensures that the same person has similar attributes, but also avoids the meaningless attributes. We hence use the above loss to update the $\mathcal{O}^{S1}$ with back propagation. We denote the resulting update dCNN as $\mathcal{O}^{S2}$.

\subsection{Fine-tuning on the Combined Dataset}

The fine-tuning in previous stage produces more accurate attribute labels. We thus consider to combine the $T$ and $U$ for the final round of fine-tuning. As illustrated in Fig.~\ref{fig:framework}, in the third stage, we first predict the attribute labels for dataset $U$ with $\mathcal{O}^{S2}$. A new dataset labeled with attribute labels can hence be generated by merging $T$ and $U$. Then, we fine-tune $\mathcal{O}^{S2}$ using sigmoid cross entropy loss on the dataset $T\&U$, which outputs the final attribute detector $\mathcal{O}$.

For any test image, we can predict its $K$-dimensional attribute label with Eq.~\eqref{eq:att_detect}. In our implementation, we only select the attributes whose confidence values predicted by $\mathcal{O}$ are larger than a specified threshold as positive, where the confidence threshold is experimentally set as 0. This essentially selects more accurate attributes. Finally, $\mathcal{O}$ produces a sparse binary $K$-dimensional attribute vector. Our person ReID system uses this binary vector as feature and measures their distance with Cosine distance to identify the same person. The validity of this three-stage training procedure and the performance of selected attributes will be tested in Section~\ref{sec:experiment}.

\section{Experiments} \label{sec:experiment}
\subsection{Datasets for Training and Testing}

To conduct the first stage training, we choose the \emph{PETA}\cite{deng2014pedestrian} dataset as the training set. Each image in \emph{PETA} is labeled with 61 binary attributes and 4 multi-class attributes. The 4 multi-class attributes are \emph{footwear}, \emph{hair}, \emph{lowerbody} and \emph{upperbody}, each of which has 11 color labels including \emph{Black}, \emph{Blue}, \emph{Brown}, \emph{Green}, \emph{Grey}, \emph{Orange}, \emph{Pink}, \emph{Purple}, \emph{Red}, \emph{White}, and \emph{Yellow}, respectively. We hence expand 4 multi-class attributes into 44 binary attributes, resulting in a 105-dimensional binary attribute label.
For the second stage training, we choose the \emph{MOT challenge}\cite{leal2015motchallenge} dataset to fine-tune dCNN $\mathcal{O}^{S1}$ with attributes triplet loss. \emph{MOT challenge} is a dataset designed for multi-target tracking and provides the trajectories of each person. We thus could get the bounding box and ID label of each person. And we use more than 20,000 images on \emph{MOT challenge}.

To evaluate our model, we choose \emph{VIPeR}~\cite{gray2007evaluating}, \emph{PRID}~\cite{hirzer2011person}, \emph{GRID}\cite{loy2013person}, and \emph{Market}~\cite{zheng2015scalable} as test sets. Note that, \emph{VIPeR}, \emph{GRID} and \emph{PRID} are included in the \emph{PETA} dataset. When we test our algorithm on them, they will be excluded from the training set. For example, when we use the \emph{VIPeR} for person ReID test, none of its images will be used for dCNN training. We do not use the \emph{CUHK} for testing, because it takes nearly one third of images in \emph{PETA}. If it is excluded, the samples for dCNN training will be insufficient.

\subsection{Implementation Details}
We select AlexNet~\cite{Krizhevsky:NIPS12} as our base dCNN architecture. We use the same kernel and filter sizes for all the hidden layers. For the loss layers of our first stage dCNN $\mathcal{O}^{S1}$ and third stage dCNN $\mathcal{O}$, we use the sigmoid cross-entropy loss layer, because each input sample has multiple positive attribute labels. We learn 105 binary attributes from \emph{PETA}. When we fine-tune dCNN with attributes triplet loss, we follow the standard triplet loss algorithm \cite{schroff2015facenet} to select samples. First randomly select the $anchor$ samples $u_{(a)}$. Then, we select samples with
 the same person ID with $u_{(a)}$ but substantially different attribute labels as $positive$ samples $u_{(p)}$. Samples from other persons having similar attribute labels with $u_{(a)}$ are selected as $negative$ samples $u_{(n)}$. Since each person only has 15 out of 105 positive attributes in average on training datasets, We select $p=10$ attributes only for initialization in Stage 2, because they can be predicted with higher accuracy, i.e., $15*60\% (the average of classification accuracy for testing) =9$.
Moreover, we select $O=0$ to ensure most testing images include near 15 positive attributes.
Parameters for learning are empirically set via cross-validation. The $\theta $ and $\gamma$ in Eq.~\ref{eq:ourloss} are set as 1 and 0.01, respectively. We implement our approach with GTX TITAN X GPU, Intel i7 CPU, and 32GB memory.

\subsection{Accuracy of Predicted Attributes}

In the first experiment, we test the accuracy of predicted attributes on three datasets, \emph{VIPeR}, \emph{PRID} and \emph{GRID}, as well as show the effects of combining different training stages. For any input image of a person, if its GroundTruth has $n$ positive attributes, we compare the top $n$ predicted attributes against the GroundTruth to compute the classification accuracy. The results are summarized in Fig.~\ref{fig:acc}. $Stage_1$ denotes the baseline dCNN $\mathcal{O}^{S1}$. $Stage_{1\&3}$ first labels $U$ with $\mathcal{O}^{S1}$, then combines $U$ and $T$ to fine-tune the $\mathcal{O}^{S1}$. $Stage_{1\&2}$ denotes the updated dCNN $\mathcal{O}^{S2}$ after the second stage training. SSDAL denotes our final dCNN after the third stage training. From the experimental results, we can draw the following conclusions:
\begin{figure}[t]
  \centering
  \includegraphics[width=0.7\linewidth]{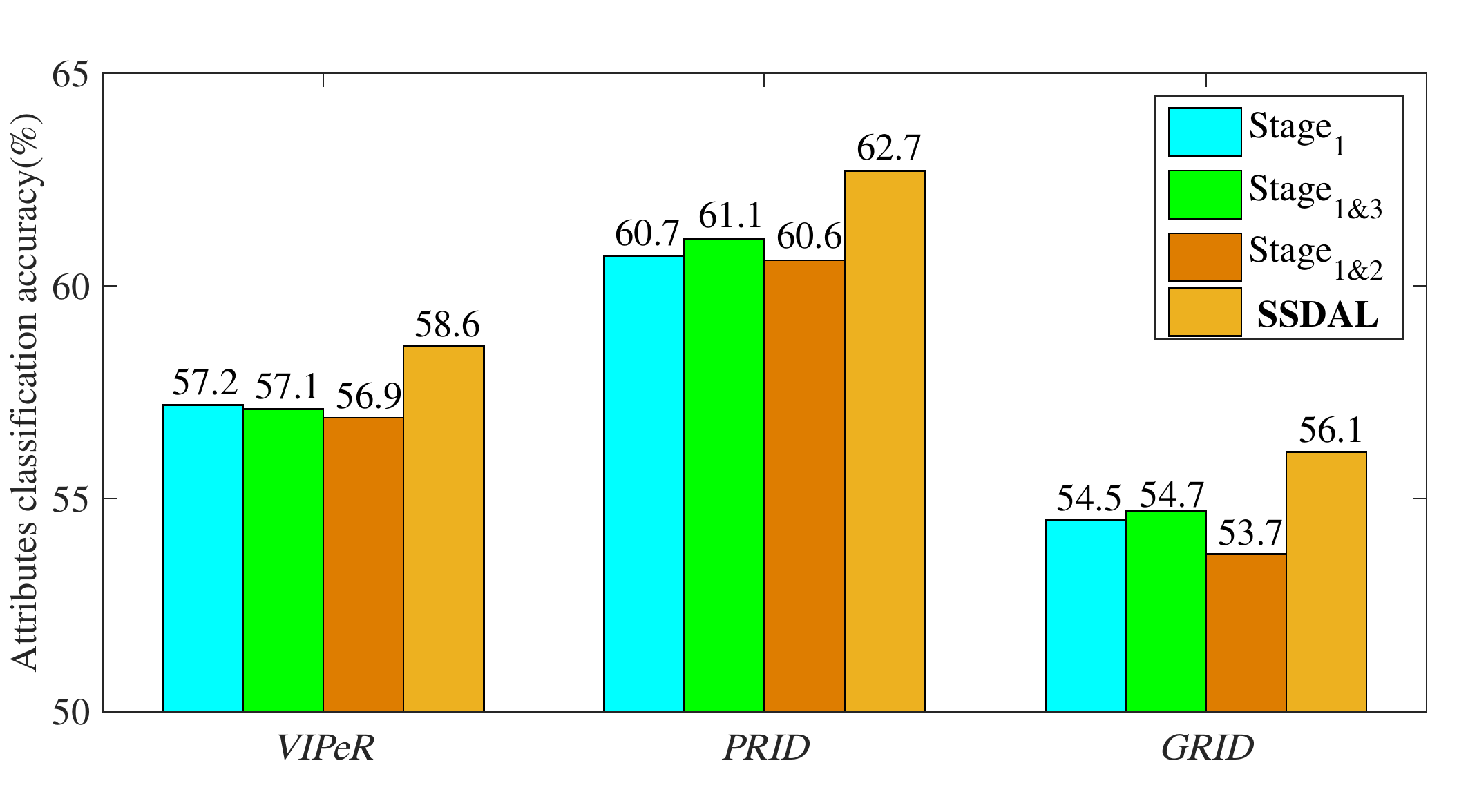}\\

  \caption{Attributes classification accuracy(\%) on three datasets.}

  \label{fig:acc}

\end{figure}

1) Although $Stage_{1\&3}$ uses larger training set, it does not constantly outperform the baseline. This is because the expanded training data is labeled by $\mathcal{O}^{S1}$, and it does not provide new cues for fine-tuning $\mathcal{O}^{S1}$ in stage-3.

2) $\mathcal{O}^{S2}$ produced by $Stage_{1\&2}$ does not constantly outperform baseline. This maybe because the weak person ID labels. Also, only updating the easily over-fitted fully-connected layers with triplet loss may degrade the generalization ability of $\mathcal{O}^{S2}$ on other datasets besides $U$.

\begin{figure}[t]
  \centering
  % Requires \usepackage{graphicx}
  \includegraphics[width=0.75\linewidth]{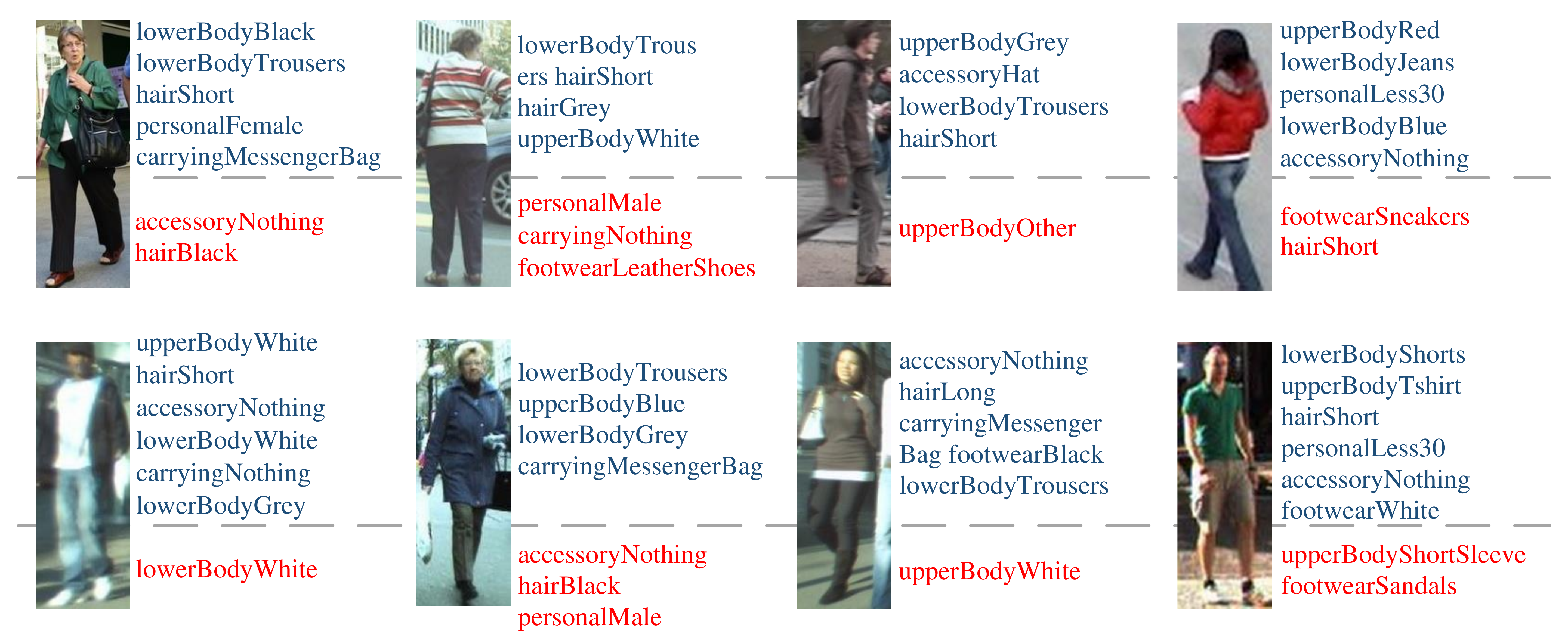}\\
  \caption{Examples of predicted attributes on \emph{MOT challenge} by the learned dCNN after three stages of training. Texts with blue color are correct attributes, while those with red color are false attributes.
}
  \label{fig:att_examples}

\end{figure}

3) SSDAL is able to improve the accuracy of baseline by $1.2\%$ in average on three datasets. This demonstrates our three-stage training framework can learn more robust semantic attributes. To intuitively show the accuracy of predicted attributes, we use the dCNN trained by SSDAL to predict attributes on \emph{MOT challenge} dataset. Some examples are illustrated in Fig.~\ref{fig:att_examples}.

\subsection{Performance on Two-Camera Datasets}
This experiment tests deep attributes on two-camera person ReID tasks. Three datasets are employed. 10 random tests are first performed for each dataset. Then, the average \emph{Cumulative Match Characteristic} (CMC) curves of these tests are calculated and used for performance evaluation. The experimental settings on three datasets are introduced as follows:
\begin{table}[t]
\tabcolsep=4pt
\caption{CMC scores, \emph{i.e.}, percentage (\%) of correct matches, of ranks 1, rank 5, rank 10, rank 20 on the \emph{VIPeR} dataset.}
{
\small
\begin{center}
\begin{tabular}{c|c|cccc}
\toprule
\multicolumn{2}{c}{Methods}      &       Rank 1 &   Rank 5 &    Rank 10 &  Rank 20  \\
\hline
\multirow{11}{3cm}{Metric\\Learning\\ based ReID}
&{ RPML}~\cite{hirzer2012relaxed} &      27.0 &     57.0 &     69.0 &     83.0  \\
&{ Salmatch}~\cite{zhao2013person}&      30.2 &     52.4 &     65.5 &     79.1 \\
  & { LMF}~\cite{zhao2014learning} &     29.1 &     52.3 &     65.9 &     80.0  \\
  &{ KISSME}~\cite{kostinger2012large} &      19.6 &     47.5 &     62.2 &     77.0  \\
&{ KCCA~\cite{lisanti:icdsc14}} &     37.3 &    71.4 &     \bf 84.6 &     92.3  \\
&{ kLFDA}~\cite{xiong2014person} &     32.2 &     65.8 &     79.7 &     90.9 \\
&{LOMO + XQDA}~\cite{liao2015person} &     40.0 &  68.9 &  81.5 & 91.1\\
&{CSL}~\cite{shen2015person} & 34.8 & 68.7 & 82.3 & 91.8\\
&{MLAPG}~\cite{liao2015efficient} & 40.7 & 69.9 & 82.3 & 92.4\\
&{TSR}~\cite{shi2015transferring} &      31.6 &     68.6 &     82.8 &     \bf 94.6  \\
&{EPKFM}~\cite{chen2015similarity} &      36.8 &     70.4 &    83.7 &     91.7  \\
\hline
\multirow{3}{3cm}{Traditional\\Attributes Learning based ReID}
&{AIR}~\cite{layne2012person} &      18.0 &     38.8 &     51.1 &     71.2  \\
&{OAR}~\cite{layne2014attributes} &     21.4 &   41.5 &     55.2  &     71.5  \\
&{LORAE}~\cite{su2015multi} &      42.3 &    \bf 72.2 &     81.6 &     89.6  \\
\hline
\multirow{3}{3cm}{Deep\\Learning\\ based ReID}
&{IDLA}~\cite{ahmed5improved} &      34.8 &     54.3 &    76.5 &     87.6  \\
&{DML}~\cite{yi2014deep} &     28.2 &     59.3 &     73.5 &     86.4  \\
&{Deep-RDC}~\cite{ding2015deep} & 40.5 & 60.8 & 70.4 & 84.4\\
\hline
\multirow{3}{3cm}{\\Proposed\\}
&{\bf $Stage_1$} & 34.5 &    63.9 &    73.1 &      87.0  \\
&{\bf SSDAL} & 37.9 &    65.5&    75.6 &     88.4  \\
&{\bf SSDAL + XQDA} & \bf 43.5 &     71.8 &     81.5 &      89.0  \\
\bottomrule
\end{tabular}
\end{center}
}

\label{table:cmcviper}

\end{table}

\textbf{\emph{VIPeR}}: 632 persons are included in the \emph{VIPeR} dataset. Two images with size $48\times128$ of each person are taken by camera A and camera B, respectively in different scenarios of illumination, postures and viewpoints. Different from most of existing algorithms, our SSDAL does not need training on the target dataset. To make fair comparison with other algorithms, we use similar settings for performance evaluation,\emph{ i.e.}, randomly selecting 10 test sets, and each contains 316 persons.

\begin{table}[t]
\caption{CMC scores, \emph{i.e.}, percentage (\%) of correct matches, of ranks 1, rank5, rank 10, rank 20 on the \emph{PRID} dataset.}
{
%\scriptsize
\small
\tabcolsep=10pt
\begin{center}
\begin{tabular}{ccccc}
\toprule
      Methods &     Rank 1 &   Rank 5 &    Rank 10 &   Rank 20  \\
\midrule
{ RPML}~\cite{hirzer2012relaxed} &     4.8 &    14.3 &    21.6 &    30.2  \\
{ PRDC}~\cite{zheng2013re} &     4.5 &    12.6 &    19.7 &    29.5  \\
{ RSVM}~\cite{prosser2010person} &     6.8 &    16.5 &    22.7 &    31.5  \\
{ Salmatch}~\cite{zhao2013person} &     4.9 &    17.5 &    26.1 &    33.9   \\
 { LMF}~\cite{zhao2014learning} &  12.5 &  23.9 &    30.7 &    36.5  \\
{ PCCA}~\cite{dikmen2011pedestrian} &     3.5 &    10.9 &    17.9 &    27.1  \\
{ KISSME}~\cite{kostinger2012large} &     4.1 &    12.8 &    21.1&    31.8 \\
{ kLFDA}~\cite{xiong2014person} &     7.6&    18.9&    25.6 &    37.4  \\
{ KCCA~\cite{lisanti:icdsc14}} &     14.5 &    34.3&   46.7  &    59.1   \\
{ LOREA}~\cite{su2015multi}  &     18.0 &   37.4 &  50.1 &  66.6  \\
{LOMO + XQDA}~\cite{liao2015person} &     15.3 &  35.7 &  41.2 & 53.8\\
{MLAPG}~\cite{liao2015efficient} & 16.6 & 33.1 & 41.4 & 52.5\\
\midrule
{\bf $Stage_1$} & 18.7&    46.9 &    55.0 &     65.8  \\
 {\bf SSDAL} & 20.1 &    47.4 &    55.7 &     68.6  \\
  {\bf SSDAL + XQDA} &  \bf 22.6 &    \bf 48.7 &    \bf 57.8 &     \bf 69.2  \\
\bottomrule
\end{tabular}
\end{center}
}
\label{table:cmcprid}

\end{table}

\textbf{\emph{PRID}}: This dataset is specially designed for person ReID in single shot. It contains two image sets containing 385 and 749 persons captured by camera A and camera B, respectively. These two datasets share 200 persons in common. For the purpose of fair comparison with other algorithms, we follow the protocol in~\cite{hirzer2011person}, and create a probe set and a gallery set, where all training samples are excluded. The probe set includes images of 100 persons from camera A. The gallery set is made up of images from 649 persons capture by camera B.

\begin{table}[t]
\tabcolsep=10pt
\caption{CMC scores, \emph{i.e.}, percentage (\%) of correct matches, of ranks 1, rank5, rank 10, rank 20 on the \emph{GRID} dataset.}
{
%\scriptsize
\small
\begin{center}
\begin{tabular}{ccccc}
\toprule
      Methods &        Rank 1 &     Rank 5 &    Rank 10 & Rank 20  \\
\midrule
{ PRDC}~\cite{zheng2013re} &      9.7 &     22.0&    33.0&     44.3 \\
{ RSVM}~\cite{prosser2010person} &     10.2 &     24.6 &     33.3 &     43.7 \\
  {MRank-PRDC}~\cite{loy2013person} &      11.1 &     26.1 &     35.8 &    46.6  \\
  {MRank-RSVM}~\cite{loy2013person} &      12.2 &     27.8 &     36.3&     49.3  \\
{RQDA}~\cite{liao2014joint} &      15.2 &     30.1&    39.2 &     49.3  \\
{EPKFM}~\cite{chen2015similarity} &     16.3 &  35.8 &  46.0 & 57.6\\
{LOMO + XQDA}~\cite{liao2015person} &     16.6 & 35.4 &  41.8 & 52.4\\
\midrule
 {\bf $Stage_1$} & 16.9 &     30.1&    40.7 &      50.2  \\
 {\bf SSDAL} & 19.1 &    35.6 &    45.8 &     58.1  \\
  {\bf SSDAL + XQDA} & \bf 22.4 &    \bf 39.2 &     \bf 48.0 &     \bf 58.4  \\
\bottomrule
\end{tabular}
\end{center}
}
\label{table:cmcgrid}
\end{table}

\textbf{\emph{GRID}}: This dataset includes images collected by 8 non-adjacent cameras fixed at a subway station. The probe set contains images of about 250 persons. The gallery set contains images of about 1025 persons, among which 775 persons do not match anyone in the probe set. For the purpose of fair comparison, images of 125 persons shared by the two sets are employed for training. The remaining 125 persons and 775 distracters are used for the testing.

\textbf{\emph{Compared Algorithms}}: We compare our approach with many recent works. Compared works that learn distance metrics for person ReID include {RPML}~\cite{hirzer2012relaxed}, {PRDC}~\cite{zheng2013re}, {RSVM}~\cite{prosser2010person}, {Salmatch}~\cite{zhao2013person}, {LMF}~\cite{zhao2014learning}, {PCCA}~\cite{dikmen2011pedestrian}, {KISSME}~\cite{kostinger2012large}, {kLFDA}~\cite{xiong2014person}, {KCCA~\cite{lisanti:icdsc14}},{TSR}~\cite{shi2015transferring}, {EPKFM}~\cite{chen2015similarity},{LOMO + XQDA}~\cite{liao2015person},{MRank-PRDC}~\cite{loy2013person}, {MRank-RSVM}~\cite{loy2013person}, {RQDA}~\cite{liao2014joint}, {MLAPG}~\cite{liao2015efficient} and {CSL}~\cite{shen2015person}. Compared works based on traditional attribute learning are {AIR}~\cite{layne2012person}, {OAR }~\cite{layne2014attributes} and {LOREA}~\cite{su2015multi}. Related works that leverage deep learning include {DML}~\cite{yi2014deep}, {IDLA}~\cite{ahmed5improved} and {Deep-RDC}~\cite{ding2015deep}. The compared CMC scores at different ranks on three datasets are shown in Table~\ref{table:cmcviper}, Table~\ref{table:cmcprid}, and Table~\ref{table:cmcgrid}, respectively.

%{RPML}~\cite{hirzer2012relaxed}, {PRDC}~\cite{zheng2013re}, {RSVM}~\cite{prosser2010person}, {Salmatch}~\cite{zhao2013person}, {LMF}~\cite{zhao2014learning}, {PCCA}~\cite{dikmen2011pedestrian}, {KISSME}~\cite{kostinger2012large}, {kLFDA}~\cite{xiong2014person}, {KCCA~\cite{lisanti:icdsc14}}, {AIR}~\cite{layne2012person},{OAR }~\cite{layne2014attributes}, {DML}~\cite{yi2014deep}, {LOREA}~\cite{su2015multi}, {TSR}~\cite{shi2015transferring}, {EPKFM}~\cite{chen2015similarity}, {IDLA}~\cite{ahmed5improved}, {LOMO + XQDA}~\cite{liao2015person}, {MRank-PRDC}~\cite{loy2013person}, {MRank-RSVM}~\cite{loy2013person}, {RQDA}~\cite{liao2014joint}, {MLAPG}~\cite{liao2015efficient}, {CSL}~\cite{shen2015person}, {Deep-RDC}~\cite{ding2015deep}.

The three tables clearly show that, even it is not fine-tuned with extra data, the baseline dCNN $\mathcal{O}^{S1}$ achieves fairly good results on three datasets, especially on \emph{PRID} and \emph{GRID}. Additionally, if we fine-tune the baseline dCNN using our attributes triplet loss, we achieve an additional $3.4\%$ improvement at rank 1 on \emph{VIPeR}, $1.4\%$ on \emph{PRID}, and $5.3\%$ on \emph{GRID}, respectively. This indicates that our three-stage training framework improves the performance by progressively adding more information into the training procedure.

Our SSDAL algorithm has surpassed all existing algorithms on the \emph{PRID} and \emph{GRID} datasets. Some recent works like {AIR}~\cite{layne2012person}, {OAR }~\cite{layne2014attributes}, and {LOREA}~\cite{su2015multi} also learn attributes for person ReID. The comparison in Table~\ref{table:cmcviper} clearly shows the advantages of our deep model in attribute prediction. Some previous works like {DML}~\cite{yi2014deep}, {IDLA}~\cite{ahmed5improved} and {Deep-RDC}~\cite{ding2015deep} take advantages of deep learning in person ReID. Different from them, our work generates camera-independent mid-level attributes, which can be used as discriminative features for identifying persons on different datasets. The experiments results in Table~\ref{table:cmcviper} also show that our method outperforms these works.

Because we use the predicted binary attributes as features for person ReID, we can also learn a distance metric to further improve the ReID accuracy. We select XQDA~\cite{liao2015person} for the distance metric learning. As can be seen from three tables, our approach with XQDA~\cite{liao2015person}, \emph{i.e.}, SSDAL + XQDA, achieves the best accuracy at rank 1 on all the three datasets. It also constantly outperforms all the other algorithms at various ranks on \emph{PRID} and \emph{GRID}. This clearly proves that our work can easily combine with existing distance metric learning works to further boost the performance.
 %This clearly indicates the validity of our proposed algorithms for person ReID.

\subsection{Performance on Multi-Camera Dataset}

 We further test our approach in a more challenging multi-camera person ReID task. We employ the \emph{Market} dataset~\cite{zheng2015scalable}, where more than 25,000 images of 1501 labeled persons are collected from 6 cameras. Each person has 17 images in average, which show substantially different appearances due to variances of viewpoints, illumination, backgrounds, \emph{etc}. This dataset is also larger than most of existing person ReID datasets. Because \emph{Market} has clearly provided the training set, we use images in the training set and their person ID labels to fine-tune our dCNN $\mathcal{O}^{S2}$.

\begin{figure}[t]
  \centering
  \begin{tabular}{c@{}c@{}}
  \includegraphics[width=.40\linewidth]{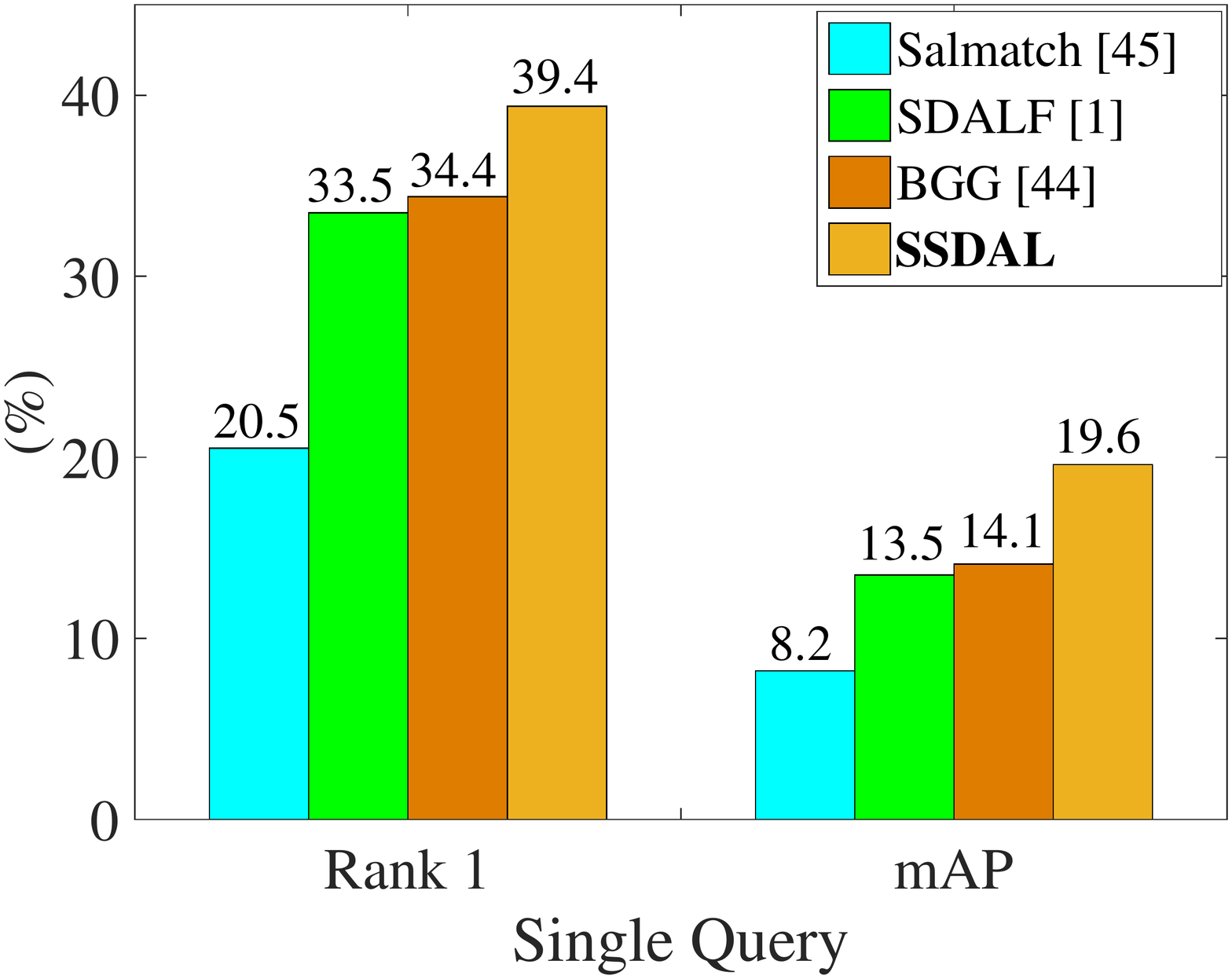}
  & \hspace{2mm}
  \includegraphics[width=.40\linewidth]{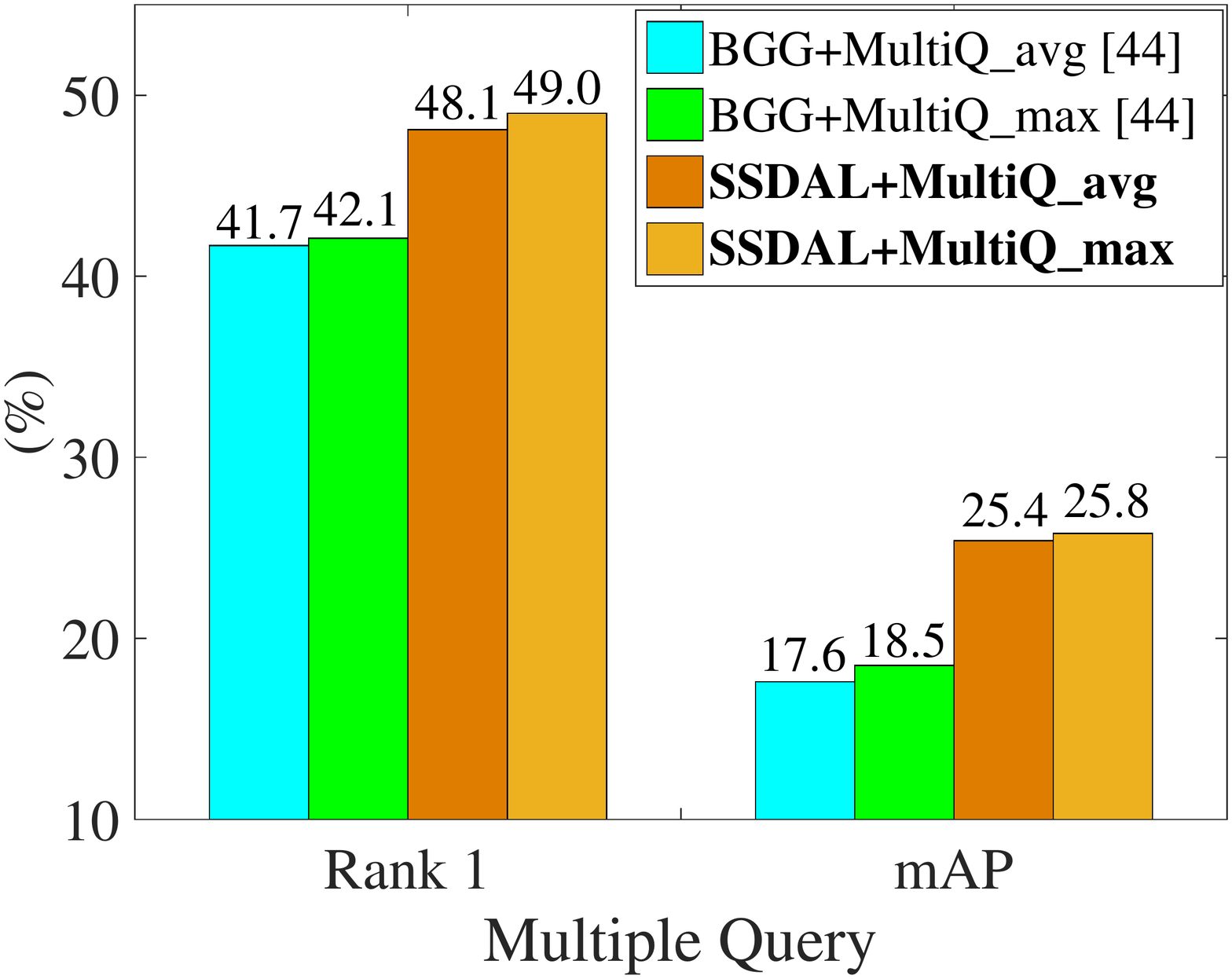}
  \end{tabular}
  \caption{CMC scores of rank 1 (Rank 1) and mean Average Precision (mAP) on the \emph{Market} dataset for the scenarios of Single Query and Multiple Query. }
  \label{fig:multicam}
\end{figure}

In contrast to the two-camera person ReID task, the multi-camera person ReID targets to identify the query person across image sets from multiple cameras. Therefore, our task is to query and rank all images from these cameras, according to the given probe image (\emph{i.e.}, Single Query) or tracklet (\emph{i.e.}, Multiple Query) of a person. Because this process is similar to image retrieval, we evaluate the performance by mean Average Precision (mAP) and accuracy at Rank 1, following the protocol in~\cite{zheng2015scalable}. The results are shown in Fig.~\ref{fig:multicam}. MultiQ\_avg and MultiQ\_max denote applying average and max pooling to acquire the final feature for a person's tracklet. More details about feature pooling can be found in~\cite{zheng2015scalable}.

From Fig.~\ref{fig:multicam}, we can observe that our approach outperforms all the compared methods by a large margin for both single query and multi-query scenarios. For the multiple query scenario, our method successfully boosts the mAP from 18.5\% to 25.8\%, resulting in an 7.3\% absolute improvement. This indicates that our method is also superior to other methods in more challenging multi-camera person ReID tasks. This experiment also shows that our learned deep attributes are robust to significant appearance variations among multiple cameras.

\subsection{Discussions}

\begin{table}[t]
\tabcolsep=10pt
\setlength{\abovecaptionskip}{0pt}
\centering
\caption{Additional experimental results on \emph{VIPeR}.}
{
%\scriptsize
%\small
\begin{center}
\begin{tabular}{ccccc}
\toprule
      Method & Rank 1 & Rank 5 & Rank 10 & Rank 20  \\
\midrule
SSDAL & 37.9 &     65.5 &     75.6 &     85.4  \\
LOMO + SSDAL + XQDA&     45.3  &  74.4 &   85.4   &   94.6  \\
  % DML &     28.2 &     59.3 &     73.5 &     86.4  \\
%\hline
%   IDLA &     34.8 &     54.3 &     76.5 &     87.6  \\
%\hline
%   AIR &      18.0 &     38.8 &     51.1 &     71.2  \\
%\hline
% OAR&     21.4 &   41.5 &     55.2  &     71.5  \\
%\hline
%\hline
FC-7 fine-tuned on $T$  &   26.5    &   48.2  &  61.1  &    72.3 \\
FC-7 fine-tuned on $U$  &     10.1  & 21.6     &  31.7    &    45.3   \\
FC-7 fine-tuned on $T+U$  &    27.4   &   49.7   &  62.3   &   74.4  \\
\bottomrule
\end{tabular}
\end{center}
}
\label{table:addresult}
\end{table}

In this part, we further discuss some interesting aspects of our method that may have been missed in the above experimental evaluations.

By using attributes features of only 105 dimensions, our method achieves promising performance on four public datasets. It is interesting to see the ReID performance after combining the compact attribute features and classic visual features. To verify this point, we integrate the appearance-based features with attributes features for better discriminative power. Table~\ref{table:addresult} shows the performance of fusing deep attributes with appearance-based feature LOMO~\cite{liao2015person}, \emph{i.e.}, LOMO + XQDA + SSDAL. It is obvious that fusing appearance-based features further improves SSDAL, \emph{e.g.}, CMC score achieves 45.3 at Rank-1. Therefore, combining with visual feature would further ensure the performance of attributes features in real applications.

%On four public datasets, the 105 predicted attributes have shown competitive performance. However, in real applications, the dimensionality of attribute feature might be too low to distinguish large-scale person images. To be applied in real applications, the learned attributes can easily integrate with appearance-based features for better discriminative power. Table~\ref{table:addresult} shows the performance of fusing deep attributes with appearance-based feature LOMO~\cite{liao2015person}, \emph{i.e.}, LOMO + SSDAL + XQDA. It is obvious that fusing appearance-based features further improves SSDAL, \emph{e.g.}, CMC score achieves 45.3 at Rank-1.

Many image retrieval works use the output of FC-7 layer in AlexNet as image feature. Therefore, another way of learning mid-level feature for person ReID is fine-tunning the FC-7 layer with triplet loss similar to the one in SSDAL, \emph{i.e.}, updating the dCNN to make same person have similar FC-7 layer features and vice versa. The FC-7 features learned in this way are also not limited to the 105 dimensions, thus might be more discriminative than attributes. To test the validity of this strategy, we fine-tune the FC-7 layer of AlexNet using person ID labels on different datasets, \emph{i.e.}, $T$, $U$, and $T+U$, respectively. Experimental results in Table~\ref{table:addresult} clearly indicates that that deep attributes outperforms such FC7 features. This clearly validates the contribution and importance of attributes.

\section{Conclusions and Future Work} \label{sec:conclusions}

In this paper, we address the person ReID problem using deeply learned human attribute features. We propose a novel Semi-supervised Deep Attribute Learning(SSDAL) algorithm. With our attributes triplet loss, images only with person ID labels can be used for training attribute detectors in a dCNN framework. Extensive experiments on four benchmark datasets demonstrate that our method is robust in attribute detection and substantially outperforms previous person ReID methods. In addition, our algorithm does not need further training on the target datasets. This means we can train the attribute prediction dCNN model only for one time, and it would work for person ReID tasks on different datasets. The dCNN model fine-tuning only requires images with person ID labels, which can be easily obtained by Multi-target Tracking algorithms. Considering the spatial locations and correlations of attributes might further improve the accuracy of attribute detection. These would be our future work.

\section{Acknowledgements} \label{sec:Acknowledgements}

This work was supported in part to Dr. Qi Tian by ARO grants W911NF-15-1-0290 and Faculty Research Gift Awards by NEC Laboratories of America and Blippar. This work was supported in part by National Science Foundation of China (NSFC) 61429201 and 61303178. This work was supported in part to Dr. Shiliang Zhang by National Science Foundation of China (NSFC) 61572050 and 91538111.

\bibliographystyle{splncs}
\bibliography{egbib}

\begin{thebibliography}{10}

\bibitem{farenzena2010person}
Farenzena, M., Bazzani, L., Perina, A., Murino, V., Cristani, M.:
\newblock Person re-identification by symmetry-driven accumulation of local
  features.
\newblock In: CVPR. (2010)

\bibitem{cheng2011custom}
Cheng, D.S., Cristani, M., Stoppa, M., Bazzani, L., Murino, V.:
\newblock Custom pictorial structures for re-identification.
\newblock In: BMVC. (2011)

\bibitem{ma2012bicov}
Ma, B., Su, Y., Jurie, F.:
\newblock Bicov: a novel image representation for person re-identification and
  face verification.
\newblock In: BMVC. (2012)

\bibitem{liu2012person}
Liu, C., Gong, S., Loy, C.C., Lin, X.:
\newblock Person re-identification: what features are important?
\newblock In: ECCV. (2012)

\bibitem{zhao2013unsupervised}
Zhao, R., Ouyang, W., Wang, X.:
\newblock Unsupervised salience learning for person re-identification.
\newblock In: CVPR. (2013)

\bibitem{wang2014person}
Wang, X., Zhao, R.:
\newblock Person re-identification: System design and evaluation overview.
\newblock In: Person Re-Identification.
\newblock (2014)  351--370

\bibitem{Zheng_2015_CVPR}
Zheng, L., Wang, S., Tian, L., He, F., Liu, Z., Tian, Q.:
\newblock Query-adaptive late fusion for image search and person
  re-identification.
\newblock In: CVPR. (2015)

\bibitem{ma2013domain}
Ma, A.J., Yuen, P.C., Li, J.:
\newblock Domain transfer support vector ranking for person re-identification
  without target camera label information.
\newblock In: ICCV. (2013)

\bibitem{dikmen2011pedestrian}
Dikmen, M., Akbas, E., Huang, T.S., Ahuja, N.:
\newblock Pedestrian recognition with a learned metric.
\newblock In: ACCV. (2011)

\bibitem{hirzer2012relaxed}
Hirzer, M., Roth, P.M., K{\"o}stinger, M., Bischof, H.:
\newblock Relaxed pairwise learned metric for person re-identification.
\newblock In: ECCV.
\newblock (2012)

\bibitem{pedagadi2013local}
Pedagadi, S., Orwell, J., Velastin, S., Boghossian, B.:
\newblock Local fisher discriminant analysis for pedestrian re-identification.
\newblock In: CVPR. (2013)

\bibitem{yan2007graph}
Yan, S., Xu, D., Zhang, B., Zhang, H.J., Yang, Q., Lin, S.:
\newblock Graph embedding and extensions: A general framework for
  dimensionality reduction.
\newblock In: PAMI. Volume~29. (2007)  40--51

\bibitem{kostinger2012large}
K{\"o}stinger, M., Hirzer, M., Wohlhart, P., Roth, P.M., Bischof, H.:
\newblock Large scale metric learning from equivalence constraints.
\newblock CVPR (2012)

\bibitem{xiong2014person}
Xiong, F., Gou, M., Camps, O., Sznaier, M.:
\newblock Person re-identification using kernel-based metric learning methods.
\newblock In: ECCV.
\newblock (2014)

\bibitem{liu2013pop}
Liu, C., Loy, C.C., Gong, S., Wang, G.:
\newblock Pop: Person re-identification post-rank optimisation.
\newblock In: ICCV. (2013)

\bibitem{li2013learning}
Li, Z., Chang, S., Liang, F., Huang, T.S., Cao, L., Smith, J.R.:
\newblock Learning locally-adaptive decision functions for person verification.
\newblock In: CVPR. (2013)

\bibitem{zheng2013re}
Zheng, W.S., Gong, S., Xiang, T.:
\newblock Re-identification by relative distance comparison.
\newblock In: CVPR. (2013)

\bibitem{wang2014personvideo}
Wang, T., Gong, S., Zhu, X., Wang, S.:
\newblock Person re-identification by video ranking.
\newblock In: ECCV.
\newblock (2014)

\bibitem{chen2015similarity}
Chen, D., Yuan, Z., Hua, G., Zheng, N., Wang, J.:
\newblock Similarity learning on an explicit polynomial kernel feature map for
  person re-identification.
\newblock In: {CVPR}. (2015)

\bibitem{liao2015person}
Liao, S., Hu, Y., Zhu, X., Li, S.Z.:
\newblock Person re-identification by local maximal occurrence representation
  and metric learning.
\newblock In: {CVPR}. (2015)

\bibitem{chen2015mirror}
Chen, Y.C., Zheng, W.S., Lai, J.:
\newblock Mirror representation for modeling view-specific transform in person
  re-identification.
\newblock In: IJCAI. (2015)

\bibitem{shen2015person}
Shen, Y., Lin, W., Yan, J., Xu, M., Wu, J., Wang, J.:
\newblock Person re-identification with correspondence structure learning.
\newblock In: ICCV. (2015)

\bibitem{liao2015efficient}
Liao, S., Li, S.Z.:
\newblock Efficient psd constrained asymmetric metric learning for person
  re-identification.
\newblock In: ICCV. (2015)

\bibitem{ding2015deep}
Ding, S., Lin, L., Wang, G., Chao, H.:
\newblock Deep feature learning with relative distance comparison for person
  re-identification.
\newblock Pattern Recognition \textbf{48}(10) (2015)  2993--3003

\bibitem{pengunsupervised}
Peng, P., Xiang, T., Wang, Y., Pontil, M., Gong, S., Huang, T., Tian, Y.:
\newblock Unsupervised cross-dataset transfer learning for person
  re-identification.
\newblock In: CVPR. (2016)

\bibitem{gray2007evaluating}
Gray, D., Brennan, S., Tao, H.:
\newblock Evaluating appearance models for recognition, reacquisition, and
  tracking.
\newblock In: PETS. (2007)

\bibitem{hirzer2011person}
Hirzer, M., Beleznai, C., Roth, P.M., Bischof, H.:
\newblock Person re-identification by descriptive and discriminative
  classification.
\newblock In: Image Analysis.
\newblock Springer (2011)  91--102

\bibitem{loy2013person}
Loy, C.C., Liu, C., Gong, S.:
\newblock Person re-identification by manifold ranking.
\newblock (2013)

\bibitem{layne2012person}
Layne, R., Hospedales, T.M., Gong, S., Mary, Q.:
\newblock Person re-identification by attributes.
\newblock In: BMVC. (2012)

\bibitem{layne2012towards}
Layne, R., Hospedales, T.M., Gong, S.:
\newblock Towards person identification and re-identification with attributes.
\newblock In: ECCV Workshops. (2012)

\bibitem{layne2014attributes}
Layne, R., Hospedales, T.M., Gong, S.:
\newblock Attributes-based re-identification.
\newblock In: Person Re-Identification.
\newblock Springer (2014)  93--117

\bibitem{layne2014re}
Layne, R., Hospedales, T.M., Gong, S.:
\newblock Re-id: Hunting attributes in the wild.
\newblock In: BMVC.
\newblock (2014)

\bibitem{su2015tracklet}
Su, C., Yang, F., Zhang, G., Tian, Q., gao, W., Davis, L.:
\newblock Tracklet-to-tracklet person re-identification by attributes with
  discriminative latent space mapping.
\newblock In: {ICMS}. (2015)

\bibitem{su2015multi}
Su, C., Yang, F., Zhang, S., Tian, Q., Davis, L.S., Gao, W.:
\newblock Multi-task learning with low rank attribute embedding for person
  re-identification.
\newblock In: {ICCV}. (2015)

\bibitem{Krizhevsky:NIPS12}
Krizhevsky, A., Sutskever, I., Hinton, G.:
\newblock Imagenet classification with deep convolutional neural networks.
\newblock In: {NIPS}. (2012)

\bibitem{Girshick:PAMI15}
Girshick, R., Donahue, J., Darrell, T., Malik, J.:
\newblock Region-based convolutional networks for accurate object detection and
  segmentation.
\newblock In: PAMI. (2015)

\bibitem{Long:CVPR14}
Long, J., Shelhamer, E., Darrell, T.:
\newblock Fully convolutional networks for semantic segmentation.
\newblock In: CVPR. (2014)

\bibitem{shankar2015deep}
Shankar, S., Garg, V.K., Cipolla, R.:
\newblock Deep-carving: Discovering visual attributes by carving deep neural
  nets.
\newblock In: {CVPR}. (2015)

\bibitem{chen2015deep}
Chen, Q., Huang, J., Feris, R., Brown, L.M., Dong, J., Yan, S.:
\newblock Deep domain adaptation for describing people based on fine-grained
  clothing attributes.
\newblock In: {CVPR}. (2015)

\bibitem{li2014deepreid}
Li, W., Zhao, R., Xiao, T., Wang, X.:
\newblock Deepreid: Deep filter pairing neural network for person
  re-identification.
\newblock In: CVPR. (2014)

\bibitem{yi2014deep}
Yi, D., Lei, Z., Li, S.Z.:
\newblock Deep metric learning for practical person re-identification.
\newblock In: {ICPR}. (2014)

\bibitem{ahmed5improved}
Ahmed, E., Jones, M., Marks, T.K.:
\newblock An improved deep learning architecture for person re-identification.
\newblock In: {CVPR}. (2015)

\bibitem{paisitkriangkrai2015learning}
Paisitkriangkrai, S., Shen, C., Hengel, A.v.d.:
\newblock Learning to rank in person re-identification with metric ensembles.
\newblock In: {CVPR}. (2015)

\bibitem{deng2014pedestrian}
Deng, Y., Luo, P., Loy, C.C., Tang, X.:
\newblock Pedestrian attribute recognition at far distance.
\newblock In: {ACM MM}. (2014)

\bibitem{leal2015motchallenge}
Leal-Taix{\'e}, L., Milan, A., Reid, I., Roth, S., Schindler, K.:
\newblock Motchallenge 2015: Towards a benchmark for multi-target tracking.
\newblock In: arXiv preprint arXiv:1504.01942. (2015)

\bibitem{zheng2015scalable}
Zheng, L., Shen, L., Tian, L., Wang, S., Wang, J., Tian, Q.:
\newblock Scalable person re-identification: A benchmark.
\newblock In: {ICCV}. (2015)

\bibitem{schroff2015facenet}
Schroff, F., Kalenichenko, D., Philbin, J.:
\newblock Facenet: A unified embedding for face recognition and clustering.
\newblock In: CVPR. (2015)

\bibitem{zhao2013person}
Zhao, R., Ouyang, W., Wang, X.:
\newblock Person re-identification by salience matching.
\newblock In: ICCV. (2013)

\bibitem{zhao2014learning}
Zhao, R., Ouyang, W., Wang, X.:
\newblock Learning midlevel filters for person reidentification.
\newblock In: CVPR. (2014)

\bibitem{lisanti:icdsc14}
Lisanti, G., Masi, I., {Del Bimbo}, A.:
\newblock Matching people across camera views using kernel canonical
  correlation analysis.
\newblock In: ICDSC. (2014)

\bibitem{shi2015transferring}
Shi, Z., Hospedales, T.M., Xiang, T.:
\newblock Transferring a semantic representation for person re-identification
  and search.
\newblock In: {CVPR}. (2015)

\bibitem{prosser2010person}
Prosser, B., Zheng, W.S., Gong, S., Xiang, T., Mary, Q.:
\newblock Person re-identification by support vector ranking.
\newblock In: BMVC. (2010)

\bibitem{liao2014joint}
Liao, S., Hu, Y., Li, S.Z.:
\newblock Joint dimension reduction and metric learning for person
  re-identification.
\newblock In: arXiv preprint arXiv:1406.4216. (2014)

\end{thebibliography}
\end{document}